\documentclass[11pt,a4paper]{article}
\usepackage{times,latexsym}
\usepackage{url}
\usepackage[T1]{fontenc}
\usepackage{tikz}
\usepackage{booktabs}
\usetikzlibrary{trees, positioning}
\usepackage{graphicx}
\usepackage{cite}
\usepackage{caption}

\usepackage[acceptedWithA]{tacl2021v1}

\usepackage{enumerate}

\title{A Systematic Survey and Critical Review on Evaluating \\Large Language Models: Challenges, Limitations,  and Recommendations}

 \author{Md Tahmid Rahman Laskar\textsuperscript{\textdagger,\textbardbl,}\thanks{\hspace{0.115cm}Corresponding Emails: \{tahmid20, jhuang\}@yorku.ca,\ \{bari0001,srjoty\}@ntu.edu.sg}\ , Sawsan Alqahtani\textsuperscript{\textsection},  M Saiful Bari\textsuperscript{\textparagraph,\footnotemark[1]}  \\ \textbf{Mizanur Rahman}\textsuperscript{\textdagger,\textbullet}\textbf{,}  \textbf{Mohammad Abdullah Matin Khan\textsuperscript{\textdaggerdbl}}\textbf{,} \textbf{Haidar Khan\textsuperscript{\textparagraph}} \\ 
 \textbf{Israt Jahan\textsuperscript{\textdagger}}\textbf{,}  \textbf{Md Amran Hossen Bhuiyan\textsuperscript{\textdagger}}\textbf{,} 
 \textbf{Chee Wei Tan\textsuperscript{\textdaggerdbl}}\textbf{,} \textbf{Md Rizwan Parvez\textsuperscript{\textdollar}} \\ 
 \textbf{Enamul Hoque\textsuperscript{\textdagger}}\textbf{, }\textbf{Shafiq Joty\textsuperscript{\textdaggerdbl,\textdegree,\footnotemark[1]} ,} \textbf{Jimmy Xiangji Huang\textsuperscript{\textdagger,\footnotemark[1]}} \\
            \small{\textsuperscript{\textdagger}York University,
             \textsuperscript{\textsection}Princess Nourah Bint Abdulrahman University},
         \textsuperscript{\textdaggerdbl}Nanyang Technological University, 
         \\
             \small{\textsuperscript{\textparagraph}National Center for AI, Saudi Arabia,
       \textsuperscript{\textdollar}Qatar Computing Research Institute (QCRI),}
          \\
             \small{\textsuperscript{\textbardbl}Dialpad Canada Inc.,
          \textsuperscript{\textbullet}Royal Bank of Canada,    
\textsuperscript{\textdegree}Salesforce Research}
          \\ 
          } 

\begin{document}
\maketitle
\begin{abstract}

Large Language Models (LLMs) have recently gained significant attention due to their remarkable capabilities in performing diverse tasks across various domains. 
However, a thorough evaluation of these models is crucial before deploying them in real-world applications to ensure they produce reliable performance. 
Despite the well-established importance of evaluating LLMs in the community, the complexity of the evaluation process has led to varied evaluation setups, causing inconsistencies in findings and interpretations.
To address this, we systematically review the primary challenges and limitations causing these inconsistencies and unreliable evaluations in various steps of LLM evaluation. 
Based on our critical review, we present our perspectives and recommendations to ensure LLM evaluations are reproducible, reliable, and robust.

\end{abstract}


\section{Introduction}
The evolution of LLMs has transitioned from simple generative models predicting the next word to advanced systems capable of following instructions and solving complex problems \cite{zhao2023survey}. Early models like GPT \cite{radford2018improvinggpt} could generate coherent text but were limited to simple tasks, whereas instruction-tuned LLMs \cite{ouyang2022training,FLAN-t5} like ChatGPT\footnote{\url{https://openai.com/index/chatgpt/}} 
greatly enhanced their versatility and ability to execute specific commands. This shift has revolutionized the development of real-world applications powered by LLMs. 

\begin{figure*}[t!]
	\centering
		\includegraphics[width=0.99\linewidth]{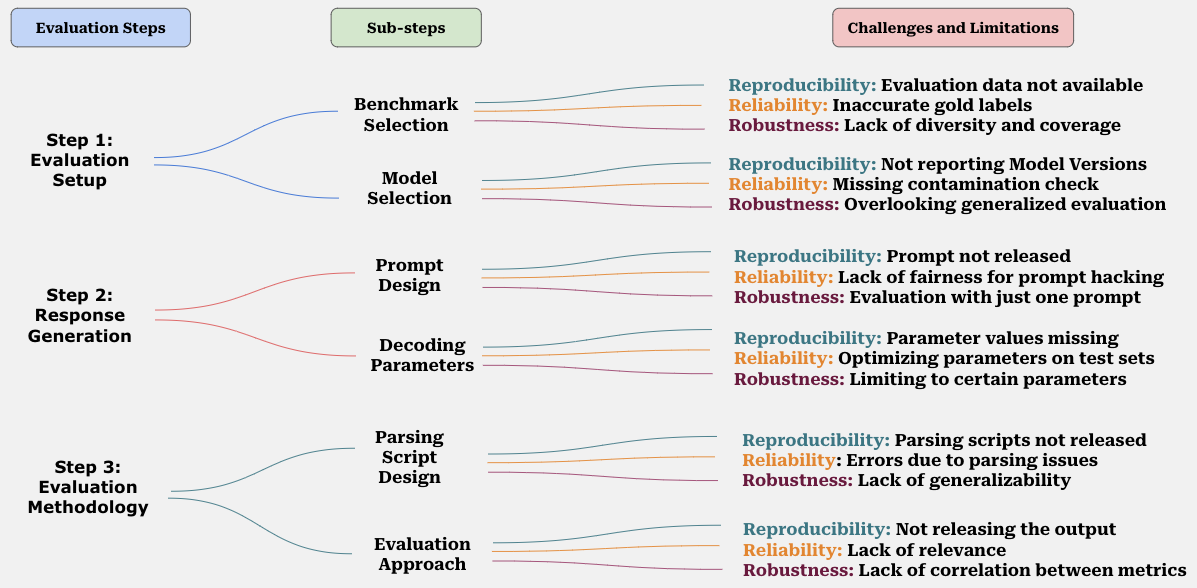}

	\caption{\small{Typology of the LLM Evaluation Workflow. \textcolor{black}{A more detailed description of the challenges and the limitations can be found in Table \ref{tab:challenges_limitations_appendix}.}} }
 \label{fig:overview}
\end{figure*}


With the advancements and broad applicability of LLMs, it is essential to properly evaluate them to ensure they are safe to use. 
This is indeed important not only for academic benchmarks but also for business use cases. Consequently, understanding the bottlenecks of current evaluation methods, 
and developing strategies to address these challenges are crucial for standardizing evaluations and enabling reliable use of LLMs in practical applications. Nonetheless, evaluating LLMs is as complex and resource-intensive as their development, involving multiple levels or aspects. 


Existing reviews \cite{llm_eval_tist,guo2023evaluating,zhuang2023through,minaee2024large,liang2022holistic} related to the evaluation of LLMs often focus only on benchmark tasks, datasets, and evaluation criteria, neglecting the broader complexities. This oversight can undermine the reliability of evaluation by ignoring issues like robustness and reproducibility. While some recent studies \cite{balloccu2024leak,mao2023gpteval} have investigated data contamination \cite{ravaut2024much} and evaluation malpractices in LLM evaluation, their focus is limited to only assessing ChatGPT, overlooking other LLMs, as well as the entire evaluation pipeline.




More recently, \citet{biderman2024lessons} discussed the reproducibility problem in existing evaluations of LLMs and introduced a library to address this. However, their work lacks comprehensive discussions on how aspects like reliability or robustness impact LLM evaluation and how to address them. Hence, 
existing LLM evaluation studies often focus on individual aspects in a scattered manner, resulting in findings that are only sparsely useful. 





To mitigate this gap, this paper brings together the discussions to address the fundamental challenges and limitations in LLM evaluations that emerge from diverse evaluation setups.
First, we craft a schematic workflow of the evaluation pipeline in practical settings (presented in Section \ref{background}) for a systematic study. We then examine each step in the evaluation workflow, uncovering various inconsistencies and decision-making complexities affecting reproducibility, reliability, and robustness (see Section \ref{validation}).
Based on our findings, we provide a principled guideline in Section \ref{opinion} to address current limitations in LLM evaluation. \textcolor{black}{The data and the code used in this paper are publicly available here: \url{https://github.com/ntunlp/Critical-Review-of-LLM-Eval}.}
\section{Overview of LLM Evaluation Process} 
\label{background}

The following components are crucial for LLM evaluation: \textit{Evaluation Setup}, \textit{Response Generation}, and \textit{Evaluation Methodology} \cite{llm_eval_tist}. Each component has its own challenges, 
which we discuss in Section \ref{validation}. These components in an evaluation workflow are shown in Figure \ref{fig:overview}.



\subsection{Evaluation Setup} 
\paragraph{Benchmark Selection:}
\label{evaluation_benchmarks}
To initiate the evaluation process of LLMs, the first step is selecting appropriate benchmarks. We categorize the benchmarking datasets into the following: \textit{general capability benchmarks},  \textit{specialized benchmarks}, and \textit{other diverse benchmarks}. 
We refer to general capability benchmarks as the ones that are often used for evaluation upon the release of an LLM (e.g., MMLU \cite{hendrycks2020measuring}, HumanEval \cite{chen2021evaluating}). In addition, there are specialized benchmarks that measure specific capabilities of LLMs (e.g., MT-Bench for chatting capabilities \cite{zheng2024judging}). There are also other benchmarks that usually combine multiple benchmarks to evaluate LLMs on diverse task (e.g.,  
HELM \cite{liang2022holistic}). 
We provide more details on each category in Appendix \ref{appendix:benchmarking_datasets}.

\paragraph{Model Selection:}\label{evaluation_model}
Selecting the appropriate model from the numerous LLMs currently available is crucial for ensuring a fair evaluation, as it helps to avoid risks such as data contamination and unfair comparisons. For a detailed discussion on prominent LLMs, see Appendix \ref{appendix:llms}.

 


\subsection{Response Generation} 
\label{evaluation_setup}
Once the benchmarks and the models are selected, the next step in the evaluation process is to {design the prompt} and {set up the decoding parameters} for response generation. In the \textbf{\textit{prompt design}} step, decisions on what type of prompting (e.g., zero-shot or few-shot) would be used are taken. 
Moreover, configuring the \textit{\textbf{decoding parameters}} (e.g., temperature) is important to ensure optimal performance \cite{shi2024thorough}. More discussions 
on this are provided in Appendix \ref{prompt_technique_appendix} and \ref{decoding_parameters_appendix}. 
\subsection{Evaluation Methodology} 
\label{evaluation_methodology}
\paragraph{Parsing Script Design:} 

Evaluating LLM-generated responses is difficult because they often produce verbose outputs (see Table \ref{tab:parsing_example} for some examples). Therefore, parsing scripts are often necessary \cite{laskar-etal-2023-systematic,jahan2024comprehensive} to extract target labels before applying evaluation metrics, ensuring alignment with evaluation criteria to maintain reliability.
\paragraph{Evaluation Approach:} 

The evaluation approach can be divided into the following: \textit{automatic evaluation}, \textit{human evaluation}, \textit{LLMs as evaluators}. 
In \textit{automatic evaluation}, before applying task-specific metrics (e.g., F1, Exact Match, Perplexity \cite{jelinek1977perplexity}), parsing scripts are often utilized to extract the targeted answer, especially in discriminative tasks. \textit{Human evaluation} 
 is required to ensure qualitative assessments of LLM responses (e.g., measuring clarity, coherence, factuality) \cite{van2021human}. Recently, human evaluation based on the Elo-based rating system \cite{zheng2024judging} has gained a lot of attention. Since human evaluation is time-consuming, 
the utilization of \textit{LLMs as evaluators} to assess other LLMs has become a popular evaluation approach \cite{huang2024empirical,chiang2023can}. More details on LLM evaluation approaches are in Appendix \ref{appendix_evaluation_approach}.

\section{Challenges in Evaluating LLMs}  
\label{validation}
We examine challenges and limitations in the evaluation process of LLMs based on three dimensions: \textit{reproducibility}, \textit{reliability}, and \textit{robustness}. 

\subsection{Reproducibility}
\label{reproducibility}
Reproducibility, the ability to consistently replicate model results under the same conditions, 
is a major challenge in generative models \cite{biderman2024lessons}. 
The primary challenge is the lack of comprehensive documentation for each part of the evaluation cycle, including benchmarking datasets, prompt construction, model details, decoding strategy, response parsing, and evaluation methodology \cite{kosch2024risk,mcintosh2024inadequacies}. Table \ref{tab:data_fairness} presents an analysis by \citet{balloccu2024leak}, revealing that a relatively low percentage of the analyzed papers shared their resources. Below, we discuss factors impacting reproducibility in the evaluation step. 

\subsubsection{Missing Details on Data \& Models Used} 


\begin{table}[t!]
    \centering
    \tiny
      \resizebox{0.45\textwidth}{!}{
    \begin{tabular}{ccccccc}
        \toprule
        \multicolumn{4}{c}{\textbf{Availability  (\%)}} &  \multicolumn{2}{c}{\textbf{Comparison  (\%)}}  \\  \cmidrule(lr){1-4} \cmidrule(lr){5-6}
         \textbf{Prompt} & \textbf{Code} & \textbf{Prompt + Code} & \textbf{Model Version} & \textbf{Fair} & \textbf{Unfair} \\
        \midrule
        90.6 & 53.3 & 50.0 & 29.3 & 20.7 & 79.3 \\
        \bottomrule
    \end{tabular}
    }
    \caption{\small{Availability of resources and fairness in model comparisons (out of 212 papers), analyzed by \citet{balloccu2024leak}.}}
    \label{tab:data_fairness}
\end{table}

\textbf{Benchmarking Data:} One factor that can negatively impede the ability to reproduce results is not releasing the exact data used for evaluation \cite{balloccu2024leak}. 
Many studies evaluate LLMs on only a subset of existing datasets \cite{bang2023multitaskchatgpt, kocon2023chatgptjackofalltrades}, while others use the exact benchmarking datasets \cite{laskar-etal-2023-systematic, qin2023chatgpt}. Despite the expectation not to compare results across studies using different subsets of the data, such comparisons often occur, as discussed by \citet{balloccu2024leak}. 
Nonetheless, without explaining the sampling strategy,
or releasing the subsets used for evaluation (and possibly their responses), reproducing results using different data subsets of the same size is challenging. 

\noindent \textbf{Model Versions:} 
The information regarding the version of a model being used is also missing in many studies \cite{biderman2024lessons,balloccu2024leak}, creating reproducibility concern (see Table \ref{tab:data_fairness}). 
The continuous updates of the closed-source models, often with undisclosed changes can also impact reproducibility. 
With these updates, earlier versions are often deprecated, and results from these versions may not apply to newer models \cite{chen2023chatgptchange}, making prior evaluation results to be no longer reproducible \cite{laskar-etal-2023-systematic, bang2023multitaskchatgpt, qin2023chatgpt, kocon2023chatgptjackofalltrades}.
Therefore, it is crucial to specify the model versions used \cite{biderman2024lessons,balloccu2024leak}, while model owners should keep earlier versions available. 

\subsubsection{Lacking Response Generation Details}

\textbf{Prompting:} 
The lack of details behind how the prompts are designed may make the findings in different literature inconsistent. For instance, variations in prompt design can lead to significantly different results, as seen in various studies \cite{jahan2024comprehensive,laskar-etal-2023-systematic,qin2023chatgpt,bang2023multitaskchatgpt}. 
While few-shot learning is found to outperform zero-shot in the original evaluation conducted by the authors of various LLMs \cite{openai2023gpt4,anil2023palm2,touvron2023llama2},  many independent evaluations demonstrate that adding few-shot examples does not necessarily outperform zero-shot models in every task \cite{ye2023comprehensive,jahan2024comprehensive}. This raises the concern of whether certain prompt engineering techniques or optimizations to select few-shot samples were applied in the original evaluations. Hence, not disclosing the details behind how the prompt is designed or how the few-shot examples are selected can hinder reproducibility.  

\noindent \textbf{Decoding Strategy:} LLMs are sensitive to decoding parameters, leading to significant performance variations based on the chosen settings \cite{roziere2023codellama, touvron2023llama2}. However, crucial details on their selection are excluded in existing literature \cite{openai2023gpt4, team2023gemini,qin2023chatgpt, bang2023multitaskchatgpt, laskar-etal-2023-systematic,kocon2023chatgptjackofalltrades}. 
This lack of transparency raises reproducibility concerns, which could be responsible for inconsistent results across studies even when similar prompts are used. For instance, \citet{qin2023chatgpt} found that adding output length restrictions in the prompt to generate summaries in no more than \textit{N} words led to a performance drop in the SAMSum dataset \cite{gliwa2019samsum}. However, \citet{laskar-etal-2023-systematic} found that such controlled experiments led to a gain in performance in the SAMSum dataset. 

\subsubsection{Evaluation Methods Unavailable}

\textbf{Parsing Scripts:} LLM-generated responses often require parsing scripts to extract desired information. 
However, as demonstrated in Table \ref{tab:data_fairness}, \citet{balloccu2024leak} observed in their analysis that almost half of the LLM evaluation papers do not release any codes. 
We also observe that most studies (these include both the LLM technical reports, as well independent evaluations) do not release their parsing scripts \cite{openai2023gpt4,team2024gemma,team2023gemini, qin2023chatgpt, bang2023multitaskchatgpt, kocon2023chatgptjackofalltrades}. Nonetheless, inaccurate design of parsing scripts may lead to different evaluation results \cite{laskar-etal-2023-systematic}. Thus, the unavailability of parsing scripts would complicate result comparisons while impacting 
reproducibility \cite{balloccu2024leak,biderman2024lessons}.
 
\noindent \textbf{Evaluation Approach:}  LLMs are increasingly used to evaluate other LLMs in development \cite{zheng2024judging}. Concerns arise due to the use of closed-source LLMs as evaluators, as their frequent updates can affect reproducibility \cite{verga2024replacingjudegjury, chen2023chatgptchange}.
Moreover, \citet{chen2023chatgptchange} observed significant behavioral changes in closed-source LLMs over short periods. 
Such reproducibility concerns are also observed in prior research that used LLMs as evaluators. 
For instance,  \citet{chiang2023can,zheng2024judging} found that using closed-source LLMs as the judge could collide with human evaluations, whereas \citet{fu-etal-2023-judge} observed the opposite. 
Since the recently proposed Prometheus-2 \cite{kim2024prometheus2}  model is an open-source alternative and demonstrates a strong correlation with humans, 
utilizing open-source LLMs as the judge 
can help mitigate the reproducibility issues prevalent with closed-source LLMs.

\subsection{Reliability} 
Reliability, the ability to trust that outcomes are as intended, is another challenge encountered during evaluation. Issues like contamination/inaccurate labels in the data, irrelevant evaluation methods, and unfair comparisons may impact the reliability of the findings, which we discuss below.  

\subsubsection{Data and Model Integrity Issues}
{\textbf{Data Integrity:}} 
Errors in benchmarks undermine accurate conclusions and model comparisons, rendering evaluations of LLMs unreliable. An integrity-compromising factor is the presence of incorrect gold labels.
For instance, existing issues in the gold labels of the widely used MMLU \cite{hendrycks2020measuring} dataset have led to the development of MMLU-Pro \cite{wang2024mmlupro} and MMLU-Redux \cite{gema2024we}. 
Recently it was also found that the coding benchmarks, HumanEval \cite{chen2021evaluating}, lacked essential test cases, leading to the development of an advanced version, HumanEvalPlus \cite{liu2024yourhumanevalplus}. 

Despite these improvements, many recent studies continue to use the older versions of datasets. For instance, despite the release of HumanEvalPlus, HumanEval is still used to benchmark LLM coding performance \cite{team2023gemini, jiang2023mistral, li2023starcoder, roziere2023codellama, gloeckle2024better, team2024gemma, Wong_2023}, potentially providing misleading insights. 
In addition, outdated labels in existing benchmarks undermine reliability of gold references. For example, in tasks like open-domain question answering, which demand real-world knowledge, many gold labels become outdated over time, as noted by \citet{laskar-etal-2023-systematic}. Consequently, even if LLMs produce correct answers, comparing them to obsolete gold labels can yield inaccurate results. Moreover, in tasks like summarization, LLM-generated summaries are often favored over human-annotated gold references \cite{pu2023summarization, zhang2024benchmarking,ding2022gptannotator}.

\noindent \textbf{Contamination in Existing Models:}
\label{contamination}
Contamination occurs when a benchmarking dataset is used in training, reducing result reliability and validity \cite{sainz2023nlpcontamination,zhou2023doncontamination,shi2023detecting}. Ensuring benchmarking examples are excluded from training data is essential to maintain reliable results. Since LLMs are pre-trained on vast amounts of text data available on the internet, this could lead to unfair evaluations if LLMs have already encountered these datasets 
during their pre-training phase \cite{ravaut2024much, xu2024benchmarking, balloccu2024leak}. 

Nonetheless, most prior LLM evaluation work focusing on zero-shot evaluation did not conduct any data contamination tests \cite{laskar-etal-2023-systematic, bang2023multitaskchatgpt, qin2023chatgpt, openai2023gpt4, team2023gemini}, raising concerns about whether these evaluations truly represent the zero-shot capabilities of LLMs. Recent research has also demonstrated a strong possibility of data contamination in many datasets used to evaluate different LLMs \cite{sainz2023nlpcontamination, li2023taskcontamination, golchin2023time, ravaut2024much, xu2024benchmarking, zhang2024carefulgsm1k, oren2023proving, balloccu2024leak, matton2024leakagecodegenerationevaluation}. With the current generation of LLMs being extremely capable of learning new skills with minimal amounts of data, exposing them to evaluation data may undermine the measurement of their true capabilities. 
Since the possibility of data contamination has led to the development of new versions of existing datasets (e.g., utilizing GSM-8K to construct GSM-1K \cite{zhang2024carefulgsm1k}), it is crucial to use  
fair 
evaluation datasets. 

\subsubsection{Lack of Fairness by Manipulating Response Generation} 


\paragraph{\textbf{Prompt Hacking:}}




One major concern in terms of lack of fairness in LLM evaluation is the possibility of prompt hacking \cite{schulhoff2023ignoreprompthacking}, which involves manipulating input prompts to a language model to elicit desired responses (e.g., biasing the outputs, or taking unfair advantages by using specific few-shot examples). 
While the performance of LLMs depends on many factors relevant to how the prompt is structured, 
most work \cite{bang2023multitaskchatgpt,qin2023chatgpt,laskar-etal-2023-systematic}, even the official technical reports  \cite{team2023gemini,openai2023gpt4,anthropicclaude3} of different LLMs lack the necessary details behind prompt construction (e.g., missing scientific validity on why a certain prompt was preferred over others, how the few-shot examples are selected, etc.). 
This makes the claims regarding the effectiveness and limitations of certain LLMs in comparison to others questionable\footnote{\url{https://crfm.stanford.edu/2024/05/01/helm-mmlu.html}}. Recognizing these parallels underscores the need for transparency and robust methodologies to ensure 
fairness in AI research and development. 

\noindent \textbf{Lack of Transparency in Decoding Parameters:} 
\citet{shi2024thorough} 
demonstrated that extensive tuning of decoding parameters could improve the performance during inference. 
However, how the different decoding parameters are selected is often underexplored in existing evaluations \cite{laskar2023building,openai2023gpt4, team2023gemini,qin2023chatgpt, bang2023multitaskchatgpt, laskar-etal-2023-systematic}, as discussed in Section \ref{reproducibility}. This poses the risk of optimizing the parameters on test sets to improve performance.

\subsubsection{Inappropriate Evaluation Methodology}
\textbf{Inaccurate Design of Parsing Scripts:}
As \citet{laskar-etal-2023-systematic} observed, evaluating LLMs entirely with an automated approach based on the answer extracted using parsing scripts may lead to an error of up to more than 10\% difference in many tasks. This raises questions about the reliability of LLM evaluations that solely depend on parsing scripts without validating the scripts' effectiveness for the task. 
To tackle this, \citet{laskar-etal-2023-systematic} proposed a hybrid approach combining parsing script-based automatic evaluation with human-in-the-loop \cite{wu2022survey,laskar2022improving}. Initially, the parsing script extracts answers from LLM-generated responses. If any issues arise, humans resolve them, enhancing the reliability of parsing-based automatic evaluation. 

\begin{figure}[t!]
	\centering
		\includegraphics[width=7cm,height=5cm]{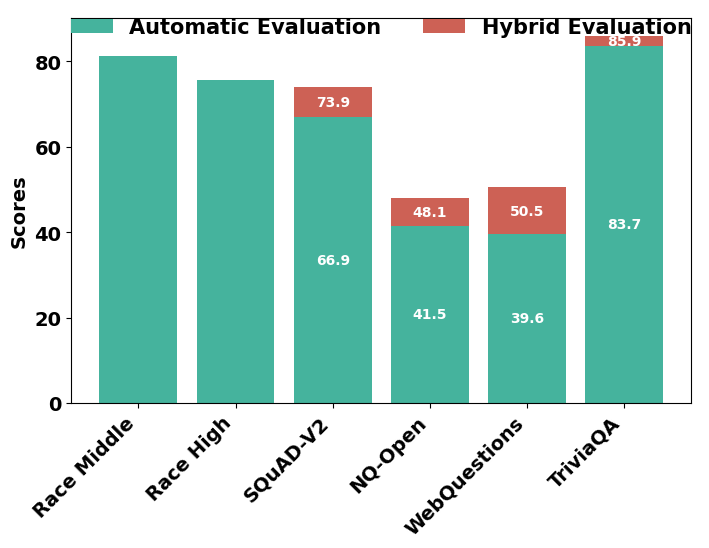}
	\caption{\small{Comparing Automatic and Hybrid Evaluation.}}
 \label{fig:parsing}
\end{figure}

In Figure \ref{fig:parsing}, we demonstrate the differences between automatic and hybrid evaluation in Open-Domain QA\footnote{NQ-Open \cite{asnq}, WebQuestions \cite{talmor2018web}, TriviaQA \cite{joshi2017triviaqa})} and reading comprehnesion datasets\footnote{SQuAD-V2 \cite{squad2}, Race-High and Race-Middle \cite{lai2017race}}.
The figure highlights the influence of human intervention on results in open-domain QA, where LLMs may generate synonymous or time-sensitive correct answers, potentially rendering gold answers outdated \cite{laskar-etal-2023-systematic}. Parsing script-based automatic evaluation is found to be reliable in Race datasets for reading comprehension, whereas notable discrepancies are observed in the SQuAD-V2 dataset. Therefore, there's a need for designing dependable parsing scripts and involving humans when appropriate. 



\noindent \textbf{Evaluation Approaches Lacking Relevancy:} In \textit{generative tasks}, utilizing automatic string-based matching techniques may not be reliable as well.  For instance, \citet{laskar-etal-2023-systematic} observed that despite LLMs scoring quite poorly on the ROUGE metric compared to SOTA summarization models, humans often prefer LLM-generated responses.
Moreover, recent research observed potential {biases} while using LLMs as evaluators, such as LLMs preferring responses generated by LLMs of the same series, positional bias \cite{stureborg2024large,wu2023style,wang2023large,bai2024benchmarking}. To mitigate this, \citet{verga2024replacingjudegjury} proposed a new technique that leveraged multiple LLMs as juries instead of using a single LLM as the judge. This approach demonstrates higher correlations with humans, while mitigating biases. 

\subsection{Robustness}
In the context of evaluating LLMs, robustness refers to the model's ability to maintain consistent performance across a wide range of inputs, conditions, or tasks. 
While there are many evaluation benchmarks currently available, 
existing work mostly relies on evaluating LLMs on some common benchmarks. 
This raises the question of whether the performance of LLMs in these common benchmarks 
reflects their true capabilities and limitations. In this section, we study the robustness of existing LLM evaluations.

\subsubsection{Lacking Generalized Evaluation}
\begin{figure}[t!]
	\centering
		\includegraphics[width=7cm,height=4cm]{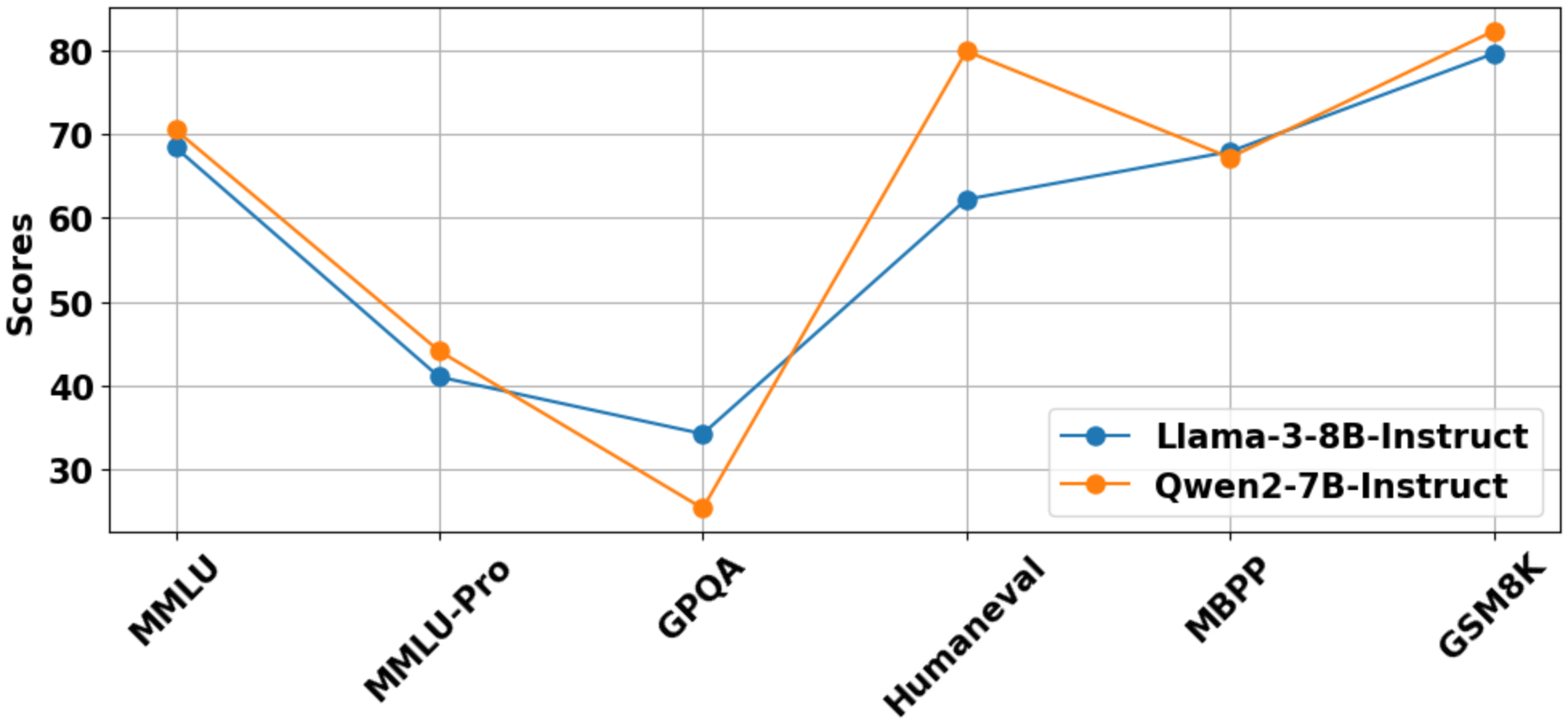}
	\caption{\small{Performance Comparison: LLaMA-3 and Qwen2}}
 \label{fig:llamavsqwen}
\end{figure}

\textbf{Limiting Evaluation to Certain Scenarios:}
Interestingly, it has been observed in recent research that certain performance gains in a specific dataset may not necessarily imply that it would also improve the performance in other datasets for similar tasks \cite{jahan2024comprehensive,sambacoe}. For instance, \citet{jahan2024comprehensive} observes that not a single LLM has superiority over other LLMs across all biomedical datasets and tasks. This is also evident if we compare the results between LLaMA-3 and Qwen2 reported in \cite{yang2024qwen2technicalreport,qwen2}. As shown in Figure \ref{fig:llamavsqwen}, while the Qwen2 model outperforms LLaMA-3 on most datasets, it falls short on GPQA and MBPP. Interestingly, for coding tasks, Qwen2 significantly outperforms LLaMA-3 on the HumanEval dataset \cite{chen2021evaluating} but not on the MBPP dataset \cite{austin2021program}. 
Meanwhile, existing common benchmarks also do not take into account some specific settings, such as how LLMs perform in long context scenarios, as recent research demonstrated that LLMs often struggle to generate the correct answer when relevant information does not appear at the beginning or end of the input context \cite{liu2024lost}. 
This highlights the importance of evaluating the generalized performance of LLMs across a set of diverse benchmarks and settings,
instead of limiting evaluation to only common benchmarks like MMLU \cite{hendrycks2020measuring}.

\begin{table}[t!]
\tiny
\setlength{\tabcolsep}{2pt} 
    \centering
    \begin{tabular}{cccccc}
       \toprule
        \multicolumn{1}{c}{{\textbf{Tokenizer
        }}} & \multicolumn{1}{c}{{\textbf{Vocab}}} & \textbf{MMLU} & \textbf{MMLU-Pro} & \textbf{MixEval} & \textbf{MixEval-Hard} \\
        \midrule
        LLaMA-2 & 32,000 & 0.52 & 0.45 & 0.29 & 0.11 \\
        LLaMA-3 & 128,256 & 0.27 & 0.21 & 0.09 & 0.03 \\
        Mistral & 32,000 & 0.59 & 0.51 & 0.31 & 0.11 \\
        Qwen2 & 151,646 & 0.22 & 0.17 & 0.08 & 0.02 \\
        \bottomrule
    \end{tabular}
    \caption{\small{Comparison of vocabulary coverage across different datasets and LLM tokenizers. The scores represent the percentage of tokenizer vocabulary that is covered by the respective dataset.}}
    \label{tab:benchmarking-datasets}
\end{table}

\noindent \textbf{Diversity and Coverage in Benchmarks:} 
Although benchmarking datasets are designed to address specific problems and objectives, 
the variation and complexity of language within these datasets are often unclear. 
\citet{liang2022holistic} highlighted 
that better coverage in benchmarking datasets would enhance the comprehensiveness of the model's evaluation. While different language models use different tokenizers to represent the benchmarking dataset, it also leads to variations in what is evaluated across models. 

As can be seen in Table \ref{tab:benchmarking-datasets}, we conducted a small-scale analysis for LLaMA-2 \cite{touvron2023llama2}, LLaMA-3,\footnote{\url{https://llama.meta.com/llama3/}} Mistral \cite{jiang2023mistral}, and Qwen2\footnote{\url{https://github.com/QwenLM/Qwen2}} on two benchmarking datasets with varying complexities: MMLU \cite{hendrycks2020measuring} and its more challenging version, MMLU-Pro \cite{wang2024mmlupro}, as well as MixEval \cite{ni2024mixeval} and its harder version, MixEval-Hard. Our findings indicate that these datasets cover a relatively small portion of the model's capabilities. Specifically, for MixEval, as the datasets became more diverse and dynamic, the vocabulary coverage for the tokenizer decreased. This trend continued as the datasets increased in difficulty, with vocabulary coverage further declining.

\subsubsection{No Tuning of Prompt and Decoding Parameters}
While various combinations of decoding parameters may lead to differences in results \cite{shi2024thorough}, possibly due to high computing requirements, existing LLM evaluation work mostly undermines the necessity of evaluating how the model performance may vary depending on its variations. Similar to the absence of decoder parameter tuning, most prior work also evaluated LLMs using only a single prompt \cite{laskar-etal-2023-systematic,qin2023chatgpt,bang2023multitaskchatgpt,kocon2023chatgptjackofalltrades,jahan2024comprehensive}. 
However, 
in the real world, users express themselves with diverse word choices, varying semantics and syntaxes, alongside minor discrepancies (e.g., misspellings or differing punctuation styles). To further examine the effects of prompt variations, we conduct an experiment using GPT-4o (2024-04-09) and GPT-3.5-Turbo (0125) \cite{openai2023gpt4}, as well as Claude-3-Opus (2024-02-29) \cite{anthropicclaude3} with the prompts used by \cite{laskar-etal-2023-systematic} and \cite{qin2023chatgpt} in the SAMSum dataset. For this experiment, the default parameters for respective LLMs are used. 

\begin{figure}[t!]
	\centering
		\includegraphics[width=7cm,height=5cm]{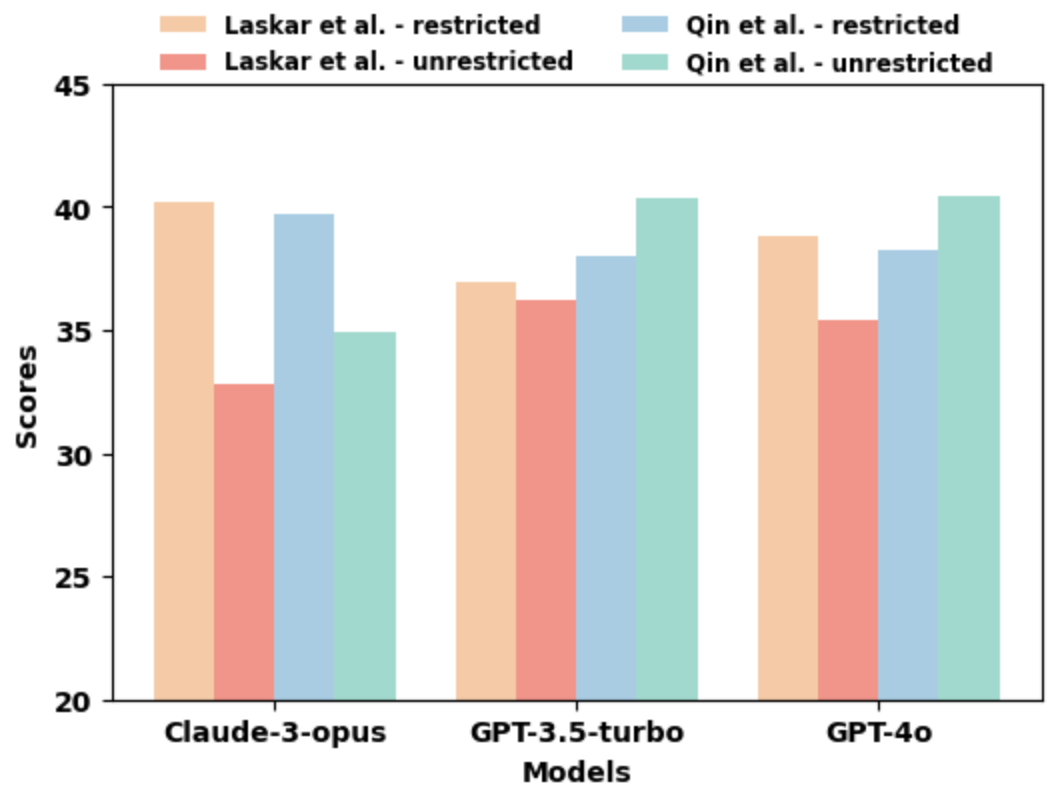}
	\caption{\small{ROUGE-1 scores in the SAMSum dataset based on Prompt Tuning.}}
 \label{fig:prompt_tuning}
\end{figure}

As shown in Figure \ref{fig:prompt_tuning}, the restricted prompting method by \citet{laskar-etal-2023-systematic} consistently outperforms the unrestricted approach across all three models. Conversely, the restricted prompting method by \citet{qin2023chatgpt} fails to surpass the unrestricted approach for GPT-3.5 and GPT-4o. However, it surprisingly outperforms the unrestricted method, indicating the significant impact of prompt tuning across models. Evaluating language models with a single prompt lacks fairness \cite{zhu2023promptbench}, yet it remains common practice \cite{bang2023multitaskchatgpt,qin2023chatgpt,laskar-etal-2023-systematic}. Minor prompt variations can lead to diverse outcomes for different models \cite{alzahrani2024benchmarks,an2023skill,zhang2024carefulgsm1k,sclar2023quantifying,lanham2023measuring,biderman2024lessons,wei2024unveiling}, highlighting the need to compare benchmarks across multiple prompts. Using automated prompt tuning techniques like Meta Probing Agents \cite{zhu2024dyval}  can ensure 
robustness to prompt variations.

\subsubsection{Evaluation Method's Generalizability and Correlation Shortcomings}
While automatic evaluations are usually utilized in discriminative tasks, they may not be applicable to every task, as demonstrated by \citet{jahan2024comprehensive} that parsing scripts are not usable in certain discriminative tasks like relation extraction. 
\citet{jahan2024comprehensive} also noted a significant performance gap between the string-matching-based ROUGE metric \cite{rouge} and the contextual similarity-based metric  BERTScore \cite{zhang2019bertscore} in text summarization. While larger models achieve better accuracy, they involve a speed-accuracy trade-off \cite{parvez-etal-2019-robust}, leading to higher costs and latency \cite{laskar2023building,fu2024tiny}. While metrics like perplexity are widely used to evaluate language models \cite{chen2023longlora}, \citet{huang2024good} found that quantized LLaMA-3 versions have lower output confidence than the original. They noted similar model rankings for perplexity and a common-sense QA dataset. However, \citet{hu2024can} found no correlation between perplexity and long context understanding tasks, highlighting the need for robust evaluations with human-correlated metrics.

This raises another question, whether automated evaluations and LLM-as-a-judge correlate with human evaluations (e.g., Elo ratings). \citet{zheng2024judging} demonstrated significant correlations between Elo ratings, LLM-as-a-judge, and automated evaluations. 
However, recent research \cite{alzahrani2024benchmarks} suggest that automated evaluations, especially those using multiple-choice questions, can yield unstable rankings with minor changes in evaluation methods. Given this instability, it prompts us to question why these automated tests should align with human Elo ratings despite demonstrating such inconsistencies. In our view, we should focus not only on correlating scores but also on how well a benchmark's rankings align with the gold standards. Analysis in Table~\ref{tab:rankings} for GPT-4 \cite{openai2023gpt4}, Gemini \cite{team2023gemini}, and Claude-3 \cite{anthropicclaude3} reveals two key observations: (i) MMLU rankings disagree with LMSys Chatbot Arena and (ii) MMLU rankings vary among themselves due to implementation differences.

\begin{table}[t!]
\centering
\tiny
\begin{tabular}{lccc}
\toprule
\textbf{} & \textbf{Chatbot} & \textbf{HELM} & \textbf{Vellum} \\ 
\textbf{Model} & \textbf{Arena} & \textbf{MMLU} & \textbf{MMLU} \\ \midrule
GPT-4o-2024-05-13 & 1 (1) & 2 (2) & 1 (1) \\ 
GPT-4-Turbo-2024-04-09	& 5 (3) & 3 (3) & 3 (3) \\  
GPT-4-0125-preview & 6 (4) & 5 (5) & 4 (4) \\  
Gemini-1.5-Pro  & 4 (2) & 4 (4) & 13 (6) \\  
Gemini-1.5-Flash & 10 (6) & 10 (6) & 10 (5) \\  
Claude-3-Opus-2024-02-29 & 7 (5) & 1 (1) & 2 (2) \\ 
\bottomrule
\end{tabular}
\caption{\small{Rankings of models on LMSys Chatbot Arena vs two MMLU implementations. The relative rank of each model in MMLU is shown in parentheses.}}
\label{tab:rankings}
\end{table}

\begin{table*}[t!]
\centering
\tiny
\begin{tabular}{p{2.25cm}p{2cm}p{4.25cm}p{4.85cm}}

\toprule
\textbf{Step} & \textbf{Sub-Step} & \textbf{Recommendation} & \textbf{Implementation: Suggested Tools or Techniques} \\ \midrule

{Evaluation Setup} & Benchmark Selection & Selected benchmarks should align with the capabilities required and updated versions of the datasets should be used to ensure reliability, diversity in the selected benchmarks is required to ensure robustness, and proper documentation of the dataset subsets is required for reproducibility. & \textbf{Reliability:} Use refined benchmarks like MMLU-Pro, Human-Eval Plus, GSM-1k to address the limitations in existing benchmarks to improve reliability. \newline \textbf{Reproducibility:} Document the data sampling technique and release the data subset used for evaluation alongside the model-generated response. \newline \textbf{Robustness:} Check tokenizer vocabulary coverage in selected benchmarks. \\ \cmidrule(l){2-4} 
& Model Selection & Data contamination check in the selected model is required for reliability, proper versioning of the model is required for reproducibility, and diverse capability evaluation (e.g., latency, memory usage, format following capability, etc.) is important to ensure robustness. & \textbf{Reliability:} Use tools like LLMSanitize Library \cite{ravaut2024much} for contamination check. \newline \textbf{Reproducibility:} Use mlflow or W\&B for documentation. \newline \textbf{Robustness:} Use tools like pyNVML to measure GPU memory requirements, FOFO for format following, compare accuracy vs latency trade-off, etc. \\ \midrule
{Response Generation} & Prompt Design & Release the prompts and few-shot examples for reproducibility, justify the selection of certain prompts and few-shot examples to ensure reliability, and compare with alternative prompts to ensure robustness. & \textbf{Reliability:} Justify the choice of certain prompts to ensure no potential of prompt hacking and compare the alternatives. Meanwhile, clearly demonstrate what and how few-shot examples are selected. \newline \textbf{Reproducibility:} Use tools like LM-Evaluation-Harness. \newline \textbf{Robustness:} Use Prompt Bench or Meta-Probing Agent. \\ \cmidrule(l){2-4} 
& Decoding Parameters & Document the decoding parameters to ensure reproducibility, justify the selection to ensure reliability, and experiment with various parameters to ensure robustness. & \textbf{Reliability:} Justify the choice of certain parameters to eliminate the risk of optimization in the test data. \newline \textbf{Reproducibility:} Use mlflow or W\&B. \newline \textbf{Robustness:} Compare the performance based on different decoding parameters, at least in a subset of the data. \\ \midrule
{Evaluation Methodology} & Parsing Script Design & Accurate parsing of the response is required for reliability, availability of these scripts is needed for reproducibility, and parsing scripts should show robustness across different models and datasets. & \textbf{Reliability:} Validate the reliability based on human evaluation, at least on a subset. \newline \textbf{Reproducibility:} Release the code. 
\newline \textbf{Robustness:} Evaluate multiple models and datasets, across all types of labels and corner cases. \\ \cmidrule(l){2-4} 
 & Evaluation Approach & Availability of the evaluation output is required for reproducibility, selected evaluation metrics should maintain correlation with humans to ensure reliability, and multiple evaluation metrics are required for evaluation robustness. & \textbf{Reliability:} Validate the effectiveness of selected metrics (e.g., measure correlation with humans), use techniques like LLM-as-juries to mitigate bias. \newline \textbf{Reproducibility:} Release the Evaluation Output. \newline \textbf{Robustness:} Use multiple evaluation metrics (e.g., in Summarization, use both word-based (e.g., ROUGE) or Contextualized (e.g., BERTScore) metrics), measure latency, GPU usage via pyNVML. \\
\bottomrule
\end{tabular}
\caption{\small{\textcolor{black}{Recommendations and Implementation Suggestions.}}}
\label{tab:rankings}
\end{table*}

\section{Recommendations and Best Practices}
\label{opinion}

So far, we've outlined the primary challenges in evaluating LLMs. In light of these challenges, a crucial question arises: How can we enhance the evaluation of LLMs? Crafting a structured framework that's both practical and easy to implement is daunting, given the complexities of generative LLM development. Previous studies tended to focus on specific evaluation aspects without offering comprehensive guidelines for the entire evaluation cycle, leaving researchers without clear guidance. Before diving into recommendations for each evaluation stage, it's important to acknowledge three key factors shaping current LLM evaluation practices: inherent randomness in generative models, significant computational demands, and insufficient documentation across stages.

\textbf{Evaluation Setup:} Selecting benchmarks for model assessment is crucial. Rather than simply replicating past choices, researchers should align datasets with required capabilities. To ensure \textit{robustness}, datasets should vary across expected LLM capabilities (e.g., long-context understanding), 
tasks (e.g., summarization), 
and language complexity (e.g., vocabulary coverage). Ideally, a metric should measure dataset diversity. For model selection, conduct contamination tests between the chosen model and benchmarks using relevant techniques \cite{ravaut2024much}. 
This acts as an additional filter for benchmarking datasets, ensuring selection of unseen ones measuring intended capabilities. Meanwhile, for \textit{reproducibility}, document any subset use of benchmarking datasets, along with 
the selected model version. In addition, throughout scientific history, intelligence progress has evolved across generations. Tests from a decade ago may appear simplistic compared to today's standards (e.g., Math Olympiads, ICPC programming contests). Refreshing LLM evaluations periodically can effectively communicate standard capabilities in both open and closed-source LLM markets and ecosystems (e.g., 
chatbots). Hence, to ensure \textit{reliability}, verify if the dataset has updated versions and incorporate them if available (e.g., HumanEvalPlus \cite{liu2024yourhumanevalplus}, 
MMLU-Pro \cite{wang2024mmlupro}, GSM-1K \cite{zhang2024carefulgsm1k}) 

\textbf{Response Generation:}
For \textit{reproducibility}, thorough documentation of prompts (e.g., explaining the selection of few-shot samples) and parameter settings (e.g., use tools like mlflow\footnote{\url{https://mlflow.org/}} or Weights \& Biases\footnote{\url{https://wandb.ai/site}} (W\&B)) is essential. To ensure \textit{reliability}, it's crucial to justify why specific prompts and parameters are chosen over others by providing comparisons with alternative options. As for \textit{robustness}, experimenting with diverse prompts and parameters is the key to showcasing their effectiveness and limitations in different scenarios. In resource-constrained environments, conducting experiments with diverse evaluation settings may pose challenges, yet it remains vital to perform robust evaluations on at least a subset of samples.


\textbf{Evaluation Methodology:}
To ensure \textit{reproducibility}, the parsing scripts and the output data used for evaluation should be published. Meanwhile, sanity-checking on the parsing script should be done to ensure \textit{reliability} and \textit{robustness} of the designed parsing script. This can be done by creating test cases for various response types, and then verifying (with human intervention if possible) whether the parsing script can reliably extract the targeted answer from the generated response. 
Meanwhile, reliance on string-based metrics like ROUGE should be minimized in favor of qualitative evaluations to ensure the \textit{reliability} of the chosen evaluation methodology. Given the cost and time constraints of human qualitative evaluation, LLM-based evaluators can be used as alternatives but must be validated for potential biases (e.g., multiple LLMs as juries instead of using a single LLM as the judge \cite{zheng2024judging}).
\textcolor{black}{Finally, \textit{robust} evaluation using task-specific metrics is encouraged. For this purpose, metrics that lack alignment with humans should be avoided. Moreover, measuring runtime latency using tools like pyNVML\footnote{\url{https://pypi.org/project/pynvml/}} 
 is recommended to evaluate the real-world applicability of different LLMs.}

\section{Conclusions and Future Work}
In this paper, we systematically survey the challenges and limitations in evaluating LLMs.
We identified significant inconsistencies and complexities at various stages of the evaluation pipeline, impacting the reproducibility, reliability, and robustness of the results. These issues underline the necessity for a standardized and systematic approach for LLM evaluation to ensure their reliable usage in real-world applications. By comprehensively reviewing the current evaluation practices, we have provided a set of recommendations aimed at enhancing the consistency and fairness of LLM evaluations. Therefore, future work should focus on developing and adopting standardized evaluation protocols for LLMs to address the identified limitations. 
This includes creating benchmark datasets, evaluation metrics, and proper documentation of the evaluation settings to ensure reproducibility, reliability, and robustness.

\section*{Acknowledgements}

We would like to thank all the anonymous reviewers for their excellent review comments. This research was supported by the Natural Sciences and Engineering Research Council (NSERC) of Canada and the York Research Chairs (YRC) program. We also acknowledge Compute Canada for the computing resources. Finally, we thank Mir Tafseer Nayeem for providing valuable feedback. 

\section*{Limitations}

One limitation of this work is that it is focused only on the evaluation phase of the LLM development cycle. Therefore, the challenges and limitations that happen during the training phase of LLMs are left out of the scope of this paper. Nonetheless, with the rapid growth of LLM technologies and huge financial incentives, it is essential to conduct a fair and reliable evaluation of LLM, alongside ensuring robustness and reproducibility, which is the focus of this work.

Another limitation of this study is that it does not study how to prevent closed-source LLMs from getting access to the online benchmarks.  
For instance, assume we have two entities: model developers and evaluators. Evaluators do not want to expose their data to the modeling team. Conversely, model developers do not want to release their model weights due to significant financial incentives. If evaluators use an API to get the responses, there is a risk that the queries may get exposed to the model developers. Therefore, without getting access to the weights, evaluators cannot reliably assess the models on their queries. Mathematically and technically, there is no fundamental way to solve this problem without altering the training dynamics which may not be an option for training teams.

Moreover, given the limited amount of study to evaluate LLMs in non-English data, our work was more focused on the monolingual scenario (mostly on English data). Therefore, investigating the challenges and limitations of LLM evaluation in multilingual and resource-constrained scenarios could be studied in the future, alongside also studying the performance of various tokenizers (both multilingual and monolingual) in LLM benchmarking \cite{rust-etal-2021-good,choo2023study}). 

Finally, the multimodal capability, in other words, the ability to understand both language and vision is another interesting capability of recently proposed LLMs~\cite{zhu2023minigpt, liu2024visual, liu2023improved, dai2024instructblip, zhang2023internlm, ye2023mplug,luo2024cheap, bai2023qwen,chen2023sharegpt4v}. 
This has led to the development of many multi-modal benchmarks \cite{fu2023mme,yu2023mm,liu2023mmbench,li2023seed,li2023seed0,liu2024visual,fu2024blink,lu2022learn,guan2023hallusionbench,li2023evaluating,qiu2024valor,chen2024we}. However, this paper was mostly focused on text-based NLP tasks and the evaluation of LLMs on multimodal benchmarks is left out for future work.

\section*{Ethics Statement}
This paper only reviews the existing challenges and limitations in LLM evaluations and provides an opinion piece and recommendation to ensure reliable, robust, and reproducible evaluations of LLMs. Thus, this review does not pose any ethical concerns.

\bibliography{tacl2021}
\bibliographystyle{acl_natbib}



\appendix
\section{Appendix}
\label{sec:appendix}

\begin{table*}[t!]
\centering
\scriptsize
\begin{tabular}{p{2cm}p{4.5cm}p{8.5cm}}

\toprule
\textbf{Criteria} & \textbf{Challenges \& Limitations} & \textbf{Description}
 \\ \midrule

{Reproducibility} & Missing Experimental Details & Lack of documentation on the data subsets used for evaluation, which few-shot examples added to the prompt, what decoding parameters are used, etc., will impact reproducibility. \\ \cmidrule(l){2-3} 
 & Not Releasing the Data & The detailed prompt as well as the response generated by the LLMs are often missing. \\ \cmidrule(l){2-3} 
 & Code Unavailable & Many studies do not release the necessary codes (e.g., parsing scripts). This may impact reproducibility of the results. \\ \cmidrule(l){2-3} 
 & Model Updates and Depreciation & Continuous updates to the closed-source models (alongside possible depreciation of the models) will create challenges for reproducing previous results. \\ \midrule
{Reliability} & Not Documenting Model Versions & The exact version of the model being used is often missing. This creates another reproducibility concern. \\ \cmidrule(l){2-3} 
 & Data Integrity & Incorrect gold labels and outdated benchmark datasets compromise evaluation reliability. \\ \cmidrule(l){2-3} 
 & Unfair Comparisons & Comparing models evaluated on the full dataset against the subset of a dataset, different few-shot examples being selected, etc. \\ \cmidrule(l){2-3} 
 & Contamination & LLMs may encounter evaluation data during pre-training, leading to contamination. \\ \cmidrule(l){2-3} 
 & Prompt Hacking & Manipulating input prompts to elicit desired responses can undermine fair evaluation. \\ \cmidrule(l){2-3} 
 & Transparency in Decoding Parameters & Lack of transparency in how decoding parameters are selected can lead to unfair comparisons. \\ \midrule
{Robustness} & Evaluation Methodology and Metrics & Reliance on string-based metrics and automated evaluation methods without proper validation can lead to unreliable results. \\ \cmidrule(l){2-3} 
 & Limiting Evaluation to Certain \newline Benchmarks & Evaluating LLMs only on a set of common benchmarks does not ensure generalizability. \\ \cmidrule(l){2-3} 
 & Lack of Diversity in Prompts \newline and Parameters & Most existing research used only a single prompt while also not tuning any of the decoding parameters, restricting the robustness of the evaluation. \\ \cmidrule(l){2-3} 
 & Insufficient Evaluation Metrics & Lack of correlation between existing evaluation metrics impacts evaluation robustness. \\ \bottomrule

\end{tabular}
\caption{\small{Challenges and Limitations in terms of Reproducibility, Reliability, and Robustness in LLM Evaluation.}}
\label{tab:challenges_limitations_appendix}
\end{table*}

\subsection{Benchmarking Datasets}
\label{appendix:benchmarking_datasets}
\paragraph{General Capability Benchmarks:} 
To benchmark the performance of LLMs, researchers typically use a set of widely recognized datasets. These common benchmarks are employed by authors upon the release of an LLM to evaluate its general capabilities. One of the most frequently used benchmarks is the MMLU benchmark \cite{hendrycks2020measuring}, which assesses LLMs' overall knowledge and reasoning abilities across various subjects. Other common benchmarks focus primarily on evaluating the commonsense reasoning capabilities of LLMs \cite{wei2022inverse}, such as HellaSwag \cite{zellers2019hellaswag}, PIQA \cite{bisk2020piqa}, SIQA, \cite{sap2019siqa}, WinoGrande \cite{sakaguchi2021winogrande}, OpenBookQA \cite{mihaylov2018openbookqa}, ARC \cite{clark2018thinkarq}. In addition, the TruthfulQA dataset  \cite{lin2021truthfulqa} is used to measure the truthfulness of an LLM, while the TyDi QA dataset \cite{mqa} is used for evaluating the information seeking question answering capability across diverse languages.  For assessing coding capabilities, the HumanEval \cite{chen2021evaluating} and the MBPP \cite{austin2021program} are two widely used benchmarks. Additional problem-solving datasets include APPS \cite{apps}, CodeContests \cite{code_contest}, and xCodeEval \cite{khan2023xcodeeval}, among others.

\paragraph{Specialized Benchmarks:}
There are also specialized benchmarks that measure specific capabilities of LLMs. For instance, the MT-Bench \cite{zheng2024judging}) evaluates whether LLMs can properly engage in conversations, the RewardBench \cite{lambert2024rewardbench} assesses the performance of reward models. Other specialized benchmarks like the AlpacaEval\footnote{\url{https://tatsu-lab.github.io/alpaca_eval/}} evaluates the instruction following capabilities \cite{zhou2023instruction} of LLMs, the Open Medical-LLM Leaderboard\footnote{\url{https://huggingface.co/blog/leaderboard-medicalllm}} evaluates the biomedical capabilities of LLMs, HHEM\footnote{\url{https://huggingface.co/spaces/vectara/leaderboard}} leaderboard for hallucination detection \cite{mishra2024fava,sadat-etal-2023-delucionqa}, BigCodeBench\footnote{\url{https://huggingface.co/blog/leaderboard-bigcodebench}} and LiveCodeBench\footnote{\url{https://huggingface.co/blog/leaderboard-livecodebench}} for code generation capability evaluation, SWE-bench \cite{jimenez2023swe} for software engineering capability evaluation. The recently proposed FOFO benchmark \citet{xia2024fofo} measures language models' ability to adhere to the requested formats in prompts across different domains. Moreover, there are also some specialized benchmarks that are used for LLM safety\footnote{\url{https://huggingface.co/spaces/AI-Secure/llm-trustworthy-leaderboard}} \cite{chao2024jailbreakbench} and red teaming\footnote{\url{https://huggingface.co/spaces/HaizeLabs/red-teaming-resistance-benchmark}} \cite{tedeschi2024alertredteaming} evaluation. The ability to understand both language and vision is another interesting capability of recently proposed LLMs~\cite{zhu2023minigpt, liu2024visual, liu2023improved, dai2024instructblip, zhang2023internlm, ye2023mplug,luo2024cheap, bai2023qwen,chen2023sharegpt4v}. 
This has led to the development of many multi-modal benchmarks \cite{fu2023mme,yu2023mm,liu2023mmbench,li2023seed,li2023seed0,liu2024visual,fu2024blink,lu2022learn,guan2023hallusionbench,li2023evaluating,qiu2024valor,chen2024we}. These benchmarks study the multimodal capabilities of LLMs across various domains, such as math and reasoning ~\cite{yue2023mmmu,lu2023mathvista}, science diagrams \cite{kembhavi2016diagram}, chart understanding and reasoning \cite{masry2022chartqa,islam2024large, masry2024chartinstruct, Rahman_2023}, document understanding \cite{mathew2021docvqa}.

\paragraph{Other Diverse Benchmarks:}
To enable a more comprehensive evaluation of LLMs across a wide range of scenarios, some studies also focused on introducing new benchmarks covering various aspects, such as HELM \cite{liang2022holistic}, PromptBench \cite{zhu2023promptbench}, OpenLLM\footnote{\url{https://huggingface.co/spaces/HuggingFaceH4/open_llm_leaderboard}}, MixEval \cite{ni2024mixeval}, etc. These benchmarks cover diverse tasks and usually include existing benchmarking datasets (e.g., MMLU, HellaSwag, BoolQ \cite{clark2019boolq}, etc.). Additionally, despite the availability of numerous benchmarks (both general and specialized), existing widely-used benchmarks still do not cover the full variety of tasks \cite{preum2018corpus,parvez2018building}. Therefore, some researchers have independently evaluated LLMs using additional diverse NLP datasets and tasks \cite{laskar-etal-2023-systematic,bang2023multitaskchatgpt,qin2023chatgpt, kocon2023chatgptjackofalltrades}. They also employed domain-specific benchmarks in fields such as biomedicine \cite{jahan-etal-2023-evaluation,jahan2024comprehensive}, finance \cite{li2023largefinance,guo2023chatgptfinance}, language-specific \cite{kabir2023benllmeval,lai2023chatgpt,khondaker2023gptaraeval,ahuja2023mega,liu2023alignbench,abdelali2024larabench}, social science \cite{ziems2024cansocialscience}, coding \cite{liu2024yourhumanevalplus}, and information retrieval \cite{zhu2023large}. In addition to that, ethics, bias, toxicity, robustness, and trustworthiness are also independently evaluated by researchers across various datasets \cite{hendrycks2020aligning,yang2022gluerobustness,wang2023robustness,zhuo2023redteaming,rawte2023surveyhallucination,liu2023mitigatinghallucination,mcintosh2024inadequacies,sun2024trustllm}. 

\subsection{Prominent LLMs}
\label{appendix:llms}
The impressive success of ChatGPT has led to the development of many LLMs in recent years. 
 Since there are hundreds of LLMs being released in recent years \cite{zhao2023survey}, we only discuss some of the prominent LLMs that achieved top rankings in various public leaderboards recently.
LLMs can be categorized into two parts: \textit{Closed-Source LLMs}: only available for use through the API or web interface, and (ii) \textit{Open-Source LLMs}: where the pre-trained weights of the model are available that allow further training of such models. Below, we present some prominent LLMs in these two categories.

\subsubsection{Closed Source LLMs} 
In the following, we categorize LLMs based on the organizations that develop these LLMs:\\

\paragraph{OpenAI models \cite{openai2023gpt4}:}

\begin{itemize}
    \item \textbf{GPT-3.5:} This model is an iteration of the GPT-3 architecture, emphasizing improvements in response quality through the application of the reinforcement learning from human feedback (RLHF) technique. GPT-3.5 is known for its robust performance in zero-shot tasks, where no specific training examples are provided during the task execution. This model has been instrumental due to its strong foundational capabilities in understanding and generating human-like text.
    \item \textbf{GPT-4:} It extends GPT-3.5's capabilities by incorporating multimodal functionalities, allowing the model to process not just text but also visual inputs. This advancement significantly broadens its applicational scope, making it adept at handling more complex tasks that require an understanding of both textual and visual information.  It features enhanced safety protocols and a sophisticated training regime that includes a safety reward signal during its reinforcement learning phase. 
    \item\textbf{GPT-4 Turbo:} This version builds upon GPT-4's foundation with substantial upgrades in computational efficiency and functionality. GPT-4 Turbo boasts an increased model capacity and an extended knowledge base that encompasses more recent data up to April 2023. It features a longer context window of up to 128,000 tokens and includes significant improvements in the model's economy and output consistency.

    \item\textbf{GPT-4o:} OpenAI's most sophisticated model, GPT-4o ("o" for "omni") is a multimodal powerhouse capable of handling both text and image inputs to generate text outputs. It improves upon GPT-4 Turbo by offering double the text generation speed and reducing operational costs by 50\%.
\end{itemize}

\paragraph{Google models:}

\begin{itemize}

\item\textbf{PaLM-2:} Released by Google in 2023, it is an advanced large language model that builds on the foundations set by its predecessor, the original PaLM. This iteration incorporates a sophisticated 'mixture of objectives' technique, allowing it to surpass the capabilities of the earlier model significantly \cite{anil2023palm2}.

\item\textbf{Gemini:} It is a multimodal model developed by google in December 2023, to understand and process a variety of information types, including text, images, audio, and video, seamlessly. Gemini's architecture allows it to perform exceptionally across multiple platforms, from large-scale data centers to mobile devices, adapting efficiently to the needs of different applications. This model sets new benchmarks in AI with its ability to excel in tasks that require complex multimodal integrations \cite{team2023gemini}.

\end{itemize}
\paragraph{Anthropic Models:} 

\textit{\textbf{The Claude series}} models, developed by Anthropic, represent a series of advanced language models designed to enhance user interaction through natural language understanding and generation. Starting with the original Claude, which excelled in tasks like summarization and creative writing, each subsequent model—Claude Instant, Claude 2.0, and the Claude 3 family (Haiku, Sonnet, and Opus)—has introduced significant improvements in processing speed, reasoning capabilities, and multimodal functionality. These models have a variety of uses, from quick response generation in Claude Instant to sophisticated multimodal understanding in Claude 3 Opus, showcasing their versatility and advanced AI technology to meet different user and enterprise needs\footnote{\url{https://www.anthropic.com/news/claude-3-family}}. The latest model in the Claude-3 series is the Claude-3.5-Sonnet\footnote{\url{https://www.anthropic.com/news/claude-3-5-sonnet}} model.

\subsubsection{Open Source LLMs} 
We similarly categorize the open-source LLMs based on the organizations that develop them:

\paragraph{Meta Models:}

\begin{itemize}
\item\textbf{Llama:} Launched in February 2023 by Meta AI, Llama was the first in the Llama series, showcasing strong performance on a range of natural language processing tasks. It competed well against larger models like GPT-3 with a smaller parameter size and was made available under a non-commercial license, primarily for academic research \cite{touvron2023llama}.

\item\textbf{Llama 2:} Released in July 2023, Llama 2 improved on its predecessor by expanding model sizes up to 70 billion parameters. It maintained the original architecture but included better training data and enhanced functionality. Notably, Llama 2 was more accessible, available for both academic and some commercial uses \cite{touvron2023llama2}.

\item\textbf{Llama 3:} In April 2024, Meta AI introduced Llama 3\footnote{\url{https://llama.meta.com/llama3/}}, the most advanced version with up to 70 billion parameters. This version added longer context capabilities and improved multimodal functions, marking a significant advancement in AI technology application across various fields.
\end{itemize}

\paragraph{Mistral Models:}

\textbf{Mistral AI}, founded in April 2023, is specialized in the development of open-source large language models. Rapidly gaining recognition in the AI industry, Mistral AI emphasizes the importance of open-source software, providing a viable alternative to proprietary models. The company has released several models, including Mistral 7B, Mixtral 8x7B, and Mixtral 8x22B, which are known for their high performance and innovation in the use of mixture of experts architectures \cite{jiang2023mistral,cai2024survey}. Codestral 22B, introduced on May 29, 2024, is a pioneering code generation model designed to enhance coding efficiency across more than 80 programming languages. With its specialized focus and lightweight architecture, Codestral significantly outperforms other leading models on the HumanEval FIM benchmark, making it a critical tool for developers seeking advanced AI-assisted coding capabilities.

\paragraph{Alibaba Models:}

\textbf{QWEN series models} are transformer-based large language models developed by Alibaba Cloud  \cite{bai2023qwen}. These models, pre-trained on diverse data sources including web texts, books, code, and more, come in various sizes ranging from 0.5 billion to 110 billion parameters. Qwen models support long context lengths and demonstrate strong performance on multiple Chinese and English evaluation tasks, including common-sense reasoning, code, and mathematics. The latest versions, Qwen 1.5 and Qwen 2, offer significant improvements in chat model performance, multilingual support, and stable support for up to 32K context length. With a comprehensive vocabulary of over 150K tokens, Qwen models are designed to handle multiple languages effectively, making them a versatile tool for various AI applications.

\paragraph{Microsoft Models:} 

The \textbf{Phi series} \cite{abdin2024phi} by Microsoft consists of small language models (SLMs) designed to provide high performance with lower computational requirements. The newly announced Phi-3 family includes models like Phi-3-mini, Phi-3-small, and Phi-3-medium, ranging from 3.8 billion to 14 billion parameters. These models excel in various benchmarks, offering capabilities similar to larger models but in a smaller, more cost-effective package. Phi-3 models are particularly suited for simpler tasks, local device operations, and environments with limited resources, making AI more accessible and efficient for diverse applications. They are available through Microsoft Azure AI Model Catalog, Hugging Face, and as NVIDIA NIM microservices. Several followup works extends Phi-models or their synthetic data into multilingual space such as \cite{boughorbel2024improving}. 

\paragraph{Technology Innovation Institute Models:} 

Technology Innovation Institute release the Falcon series models \cite{almazrouei2023falcon},  such as the Falcon 2 series that include models with parameter sizes of 1.3B, 7.5B, 40B, and 180B. These models are notable for their use of the REFINEDWEB dataset. Falcon models are designed for both research and commercial use, with Falcon 2 models featuring multilingual and multimodal capabilities, including vision-to-language. The Falcon 180B model, in particular, is accessible under a royalty-free license.

\paragraph{Cohere Models:}

Cohere offers a variety of advanced large language models designed for multiple use cases, including text generation, embeddings, and reranking. The Command family models, such as Command R+ and Command R, excel in conversational tasks and complex workflows like code generation and retrieval-augmented generation (RAG) \footnote{\url{https://cohere.com/command}} \cite{lewis2020retrieval, gao2023retrieval,chen2024benchmarking,liu2023recall,xiong2024benchmarking,tang2024multihop,alonso2024medexpqa,lyu2024crud, parvez-etal-2021-retrieval-augmented, parvez-etal-2023-retrieval, wang2023learning}. The Embed models enhance search, classification, and clustering capabilities with both English and multilingual support. The Rerank models improve search algorithms by re-organizing results based on specified parameters. Cohere models are accessible across platforms like Amazon SageMaker, Microsoft Azure, and Oracle GenAI Service, enabling seamless integration into diverse applications
 and retrieval augmented generation.

 \paragraph{Google Gemma Models:}
While early LLMs released by Google's are mostly closed-source (e.g., PalM-2, Gemini, etc.), Google has also recently released some lightweight open-source LLMs, named as Gemma\footnote{\url{https://storage.googleapis.com/deepmind-media/gemma/gemma-report.pdf}} family LLMs, that also have multimodal capabilities\footnote{\url{https://huggingface.co/blog/paligemma}}.

\subsection{Prompting Techniques} 
\label{prompt_technique_appendix}
Prompts can be designed in various ways \cite{Schulhoff2024ThePR,gpt3,wei2022chainofthought,FLAN-t5, islam2024mapcoder, parvez2024evidence}, as stated below:

\begin{itemize}
    \item \textbf{In-Context Learning (Zero-shot):} It means that the prompt used to interact with the model contains no examples or demonstrations. The model relies on its pre-existing knowledge, obtained from its initial training on diverse data, to generate a response or perform the task based solely on the instructions given. For example, “classify the sentence as biased or unbiased text”. 
\item \textbf{In-Context Learning (Few-shot):} It means that the prompt used to interact with the model includes a small number of examples or demonstrations. The model uses these examples to quickly adapt and understand how to perform a specific task, leveraging the details within these examples. This technique allows the model to extend its pre-existing knowledge to new tasks by closely analyzing the limited examples given. For instance, classify the sentence as biased or unbiased based on a few similar examples provided.
    \item \textbf{Chain-of-Thought Prompting (CoT):} This technique encourages models to generate intermediate reasoning steps before arriving at a final answer, mimicking a human-like problem-solving approach. This can be combined with few-shot prompting to achieve better results on more complex tasks. For example, if asked to determine whether the number "15" is odd or even, the model might outline its reasoning as follows: "An even number is divisible by 2 without a remainder. 15 divided by 2 is 7 with a remainder of 1. Therefore, 15 is an odd number." This step-by-step explanation helps clarify the model's thought process and supports its conclusion. 
        \item \textbf{Decomposition Techniques:} These techniques break down complex problems into simpler sub-problems that can be solved sequentially by the GenAI model. Each component of the problem is addressed individually, and the solutions are integrated to form a comprehensive response. Decomposition is especially useful in tasks that require layered reasoning or have multiple steps. For example, in solving a math word problem, decomposition might involve separately calculating the distances each person travels and then combining these calculations to determine when they meet.
        \item \textbf{ Role-based and Style-based Prompting:} In these techniques prompts are designed to induce a specific style or persona in the model's responses. By specifying a role (e.g., a scientist explaining a concept) or a style (e.g., formal or poetic), users can guide the tone and formality of the AI's output. This technique is valuable in applications requiring genre-specific content generation or when the output needs to fit a particular communicative context.
            \item \textbf{Prompt chaining:} It is a technique where a complex task is divided into simpler subtasks, each addressed by its own prompt. The response from one prompt is used as the input for the next, creating a sequential chain of prompts that gradually build towards the final answer. This method enhances the performance and reliability of large language models by breaking down tasks into manageable parts, making it easier to control and refine the model's responses at each step. For example, in a document analysis task, the first prompt might extract key facts from a text, and the second prompt would use these facts to generate a summary.
            
    \item \textbf{Tree of Thoughts (ToT):} It is a technique that structures problem-solving into a tree of possible solutions. It uses strategies like breadth-first or depth-first search to evaluate each potential solution path. For example, in solving a puzzle, ToT might explore different moves to find the quickest solution path.
    
    \item \textbf{Directional Stimulus Prompting (DSP) :} It is a technique that enhances how large language models (LLMs) respond to tasks by using dynamically generated prompts. A secondary, tuneable model creates specific hints that guide the main, unchangeable LLM to produce more targeted and relevant outputs. This method uses reinforcement learning to refine these prompts based on how well they perform, making DSP a more adaptive and precise approach compared to standard prompting techniques. For instance, in summarizing complex documents, DSP might generate a prompt like "Summarize focusing on economic impacts," guiding the LLM to tailor its output specifically to the economic aspects mentioned in the text. 
        \item \textbf{Multimodal Prompting:} Extending beyond text, multimodal prompting involves using inputs like images, audio, or video along with textual descriptions. This technique leverages the model's capability to process and integrate information from diverse data types, enhancing its applicability in scenarios where multiple forms of data are available. For example, interpret a scene from a video by analyzing both the spoken dialogue and the visual content to determine the mood of the conversation.

            \item \textbf{Meta-Prompting:} It  involves creating prompts that instruct the AI to generate or refine its prompts, essentially using AI to improve the efficiency and effectiveness of prompt engineering. This recursive use of prompting can lead to more dynamic and contextually adaptive AI behaviors. For example, ask the AI to optimize a prompt that instructs another AI to summarize news articles, thereby refining the instructions to enhance summary relevance and conciseness.

\end{itemize}

\subsection{Decoding Parameters} 
\label{decoding_parameters_appendix}
There are various decoding parameters that are required to be set. For instance:

\begin{itemize}
    \item \textbf{Temperature:} It is used to control the randomness of the output. It is typically between 0 and 1. Lower values (e.g., 0.1) make the model more deterministic and focused on the most likely next token, while higher values (e.g., 0.9) introduce more randomness and diversity.
    \item \textbf{Beam Size:} It refers to the number of beams in Beam Search \cite{freitag2017beam}, a decoding strategy that keeps track of multiple possible sequences (beams) at each step of generation to find the most likely sequence. A higher number of beams usually leads to more accurate results but at the cost of increased computation. 
    \item \textbf{Top-K:} The number of top probable tokens to consider. For example, if K=10, the model will choose the next token only from the top 10 most likely tokens.
    \item \textbf{Top-P:} The cumulative probability threshold. For example, if P=0.9, the model will sample from the smallest set of tokens whose combined probability is at least 90\%.
      \item \textbf{Maximum Output Tokens:} It sets the maximum number of tokens 
      to generate. 
    
\end{itemize}

\begin{table*}[t!]
\centering
\small
\resizebox{.975\textwidth}{!}{%
\begin{tabular}{p{\linewidth}p{\linewidth}p{\linewidth}}
\toprule
\textbf{Scenario 1:} For the response generated, designing a parsing script to extract the answer ``Lionel Messi'' is straightforward. However, the parsing script should also be robust to cover cases like abbreviations, uppercase-lowercase sensitivity, punctuations, synonyms, stemming, lemmatization, paraphrases, etc. \\
\midrule
\textbf{Prompt:} Which player has won the best player award in Fifa world cup 2022? \\
\midrule
\textbf{Sample LLM Response (GPT 4o):} Lionel Messi won the Best Player award (Golden Ball) in the FIFA World Cup 2022. He was instrumental in leading Argentina to victory in the tournament, culminating in their triumph in the final against France.\\
\midrule
\textbf{Correct Answer:} Lionel Messi \\
\toprule \toprule \toprule
\textbf{Scenario 2:} While Extraction of the answer ``Lionel Messi'' is required, due to the LLM knowledge-cut-off date of September 2021, it may answer about 2018. However, the target answer ``Lionel Messi'' is also in the output and so if the parsing script only parses the target answer then it may consider the response as correct whereas the response is wrong.\\
\midrule
\textbf{Prompt:} Which player has won the best player award in the last Fifa world cup? \\
\midrule
\textbf{Sample LLM Response (Older ChatGPT 3.5 having knowledge cut-off date of September 2021):} The Best Player award (Golden Ball) in the previous FIFA World Cup, which was held in 2018 in Russia, was won by Luka Modric from Croatia. Prior to the that, Lionel Messi had won it in 2014. \\
\midrule
\textbf{Correct Answer:} Lionel Messi \\
\bottomrule
\end{tabular}}
\caption{\small{Some examples of LLM-generated response requiring parsing script to extract the target answer. For Scenario 2, human evaluation is usually needed to ensure accurate parsing of the answer.}}
\label{tab:parsing_example}
\end{table*}

\subsection{Parsing Script Design} 

While there are various evaluation software \cite{biderman2024lessons,dalvi2023llmebench} currently available, they are limited to certain scenarios (e.g., limited to certain datasets and benchmarks, prompts, etc.). Thus, for the evaluation of LLMs across diverse settings, researchers often require to write parsing scripts. We present some scenarios in Table \ref{tab:parsing_example} to demonstrate why parsing script is required for such cases and the importance of validating parsing scripts. 


\subsection{Evaluation Approach} 
\subsubsection{Automatic Evaluation} 
\label{appendix_evaluation_approach}

To provide a high-level overview, automatic evaluation for LLMs can be divided into the following:

\textbf{\textit{Language Modeling:}} Perplexity \cite{jelinek1977perplexity} is widely used to study the performance of auto-regressive language models. It measures how confidently a model predicts the next word in a sequence, with the assumption that lower perplexity indicates better performance. Hence, perplexity has been historically used to assess the language model's capability to generate a coherent language and is also useful to quickly compare different models or checkpoints.




\textbf{\textit{Discriminative Tasks:}}
For tasks involving class prediction, post-processing using a parsing script is usually required to extract answers from the LLM-generated responses to compare against gold labels. In this context, metrics such as Exact Match, Accuracy, Precision, Recall, and F1, are usually utilized in discriminative tasks \cite{laskar-etal-2023-systematic,qin2023chatgpt,bang2023multitaskchatgpt}. Since metrics like exact match have several limitations (e.g., they do not consider the synonym of the gold label), various metrics for certain tasks (e.g., question answering \cite{chen2020mocha,bulian2022tomayto,manas2024improving,li2024pedantspreciseevaluationsdiverse}) are proposed.  

\textbf{\textit{Generative Tasks:}} For generative tasks such as summarization or machine translation, parsing scripts are usually not required \cite{laskar-etal-2023-systematic,jahan2024comprehensive} and so the full response generated by LLMs are compared against the gold reference. In this regard, \textit{ROUGE} \cite{rouge} and \textit{BLEU} \cite{bleu} which are based on n-gram word matching are widely used. Meanwhile, various contextualized similarity \cite{parvez-chang-2021-evaluating,laskar2020contextualized} metrics (e.g., \textit{BERTScore} \cite{zhang2019bertscore}, \textit{BARTScore} \cite{yuan2021bartscore}, \textit{AlignScore} \cite{zha2023alignscore,wang2024lessalignscore}) are also utilized that do not depend on word-based similarity measures.
\subsubsection{Human Evaluation}
Since LLMs generate human-like responses, it is often required to conduct qualitative evaluation of their responses. Earlier, qualitative evaluation of model-generated responses in terms of fluency, coherence, and informativeness were very popular \cite{laskar2022domain}.  However, with LLMs usually generating informative, fluent, and coherent response \cite{laskar-etal-2023-systematic,bang2023multitaskchatgpt,qin2023chatgpt,kocon2023chatgptjackofalltrades}, the evaluation of factual consistency of LLM-generated responses has become more important recently \cite{fu-etal-2023-judge}. Moreover, qualitative evaluation to compare between LLM-generated responses via leveraging humans based on the Elo rating system \cite{zheng2024judging} has gained a lot of attention. 

 \begin{figure*}[t!]
	\centering
		\includegraphics[width=\linewidth]{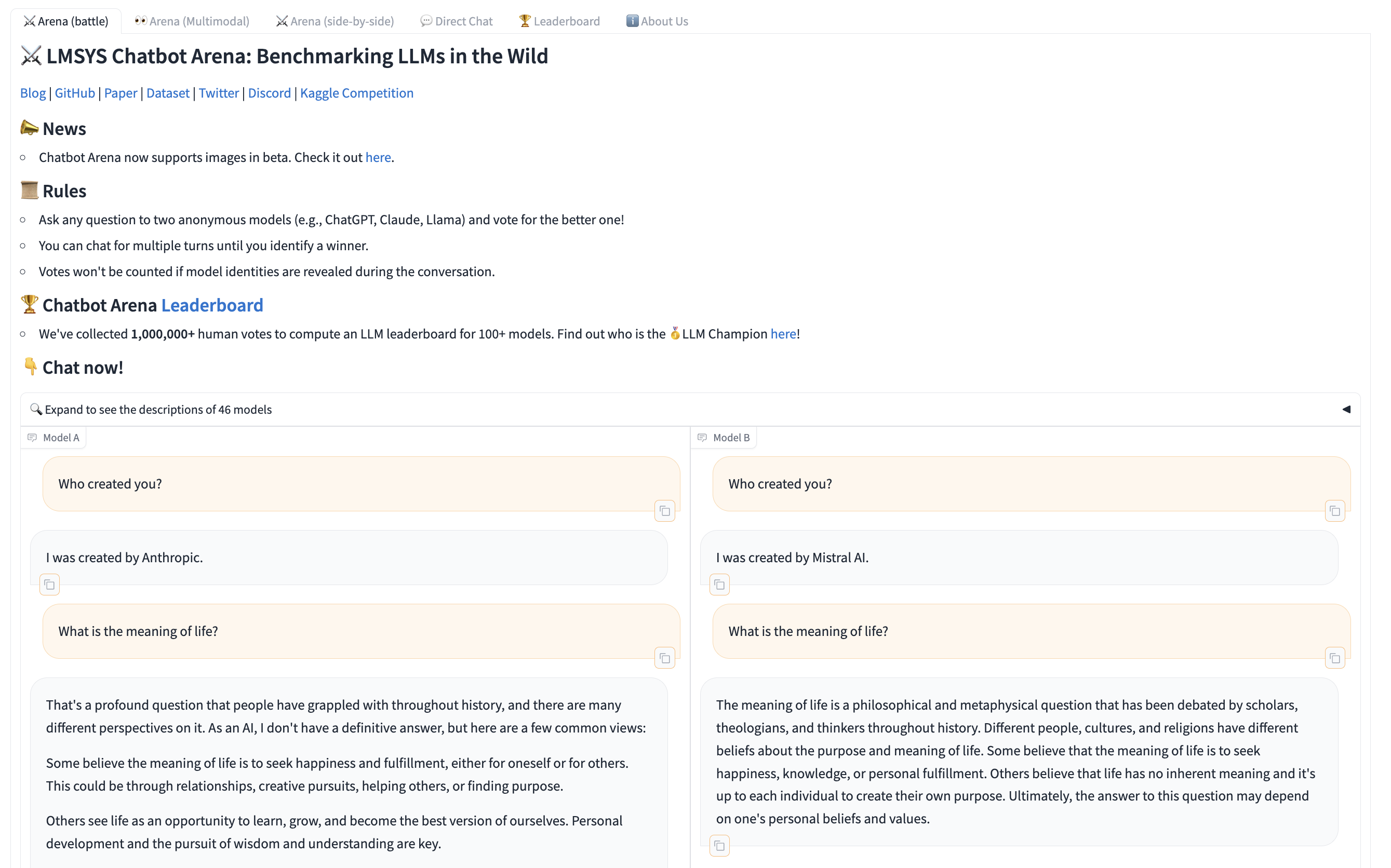}
	\caption{Ownership attack for blind evaluation on LLMs: Reviewers can pose any ownership-related questions and select their preferred model solely based on the ownership of the model. LMSys doesn't count votes if the model's identities are revealed during conversation}
 \label{fig:elo-hacking}
\end{figure*}

\paragraph{\textit{Elo Rating:}}
\label{elo_rating_system}
Elo rating works by comparing LLMs in pairwise ``A vs B'' comparisons, where each model is assigned an initial numerical rating \cite{ boubdir2023elo, zhao2023slic}. The outcome of each comparison adjusts these ratings based on the Elo algorithm: if a model performs better than expected, its rating increases; if it performs worse, its rating decreases. The expectation of a model's performance is calculated using its rating relative to its opponent's, adjusted by a factor that represents the sensitivity of expected scores to differences in ratings. To ensure a robust evaluation of LLMs using the Elo benchmark, it's important to follow key indicators like reliability and transitivity \cite{ boubdir2023elo}. Reliability keeps Elo ratings consistent across various comparison sequences and prevents them from being overly sensitive to changes in hyperparameters, such as the K-factor. Transitivity is crucial, indicating that if model A is rated higher than model B, and model B is rated higher than model C, model A should logically rank above model C. Extensive testing with both synthetic and real-world data is essential to verify that Elo scores accurately and stably reflect model performance \cite{ boubdir2023elo}. This involves making precise adjustments to the comparison order, selecting hyperparameters carefully, and utilizing numerous permutations to ensure outcome consistency. Due to the sensitive nature of the Elo rating system towards the order in which the updates were performed, \citet{zheng2024judging} used the Bradley-Terry (BTL) model for their chatbot arena ranking. It is observed that model A can have a higher win rate than model B both empirically and statistically but a lower Elo rating. Since win rate serves as the stand-in measure for the probability of a model being better than another, this signifies the findings by \citet{boubdir2023elo} that Elo rating is non-transitive with or without (BTL). On the other hand, BTL-based rating is tolerant to an imbalanced number of votes per model as shown by \cite{zheng2024judging}, they also propose a different probability of win rates that are derived from the ratings found from BTL which is transitive but doesn't correlate with the empirical win rates. 

{\textbf{\textit{Elo hacking:} }}
Crowdsourced Elo-based ranking has gained popularity through the LMSys leaderboard \footnote{\url{https://huggingface.co/spaces/lmsys/chatbot-arena-leaderboard}} and has been accepted by various organizations, prompting them to release their LLMs early into this ecosystem for human evaluation. However, such setups can be easily exploited on a large scale using simple techniques. Figure \ref{fig:elo-hacking} illustrates how someone can initially bypass the blind scoring mechanism through ownership hacking. Additionally, the evaluation of knowledge bases is not easily tracked, making votes on highly complex reasoning questions equivalent to those on simpler queries. Furthermore, upon the release of a popular model, systematic attacks or boosting can be initiated through ownership hacking. In addition to that, considering same score for \emph{tie} and \emph{both-bad} can significantly change leaderboard position. We recommend to use \emph{tie} as $0.5$ point and  \emph{both-bad} as $0$ point.

\subsubsection{LLMs as Evaluators}


Since human evaluation is time-consuming  \cite{laskar2023can,laskar2023cqsumdp} and difficult to reproduce, 
the instruction-following capabilities of LLMs have also inspired researchers to use certain LLMs as a judge to evaluate the responses generated by other LLMs \cite{perez2022redteaming,huang2024empirical,fu-etal-2023-judge,hada2023large,chern2024can,lu2024llmscore,kobayashi2024large,kocmi2023large,shankar2024validates,luo2023chatgpt,gao2023human,kim2024evallm,kenton2024scalable}. While prior work mostly utilized general-purpose closed-source LLMs-as-a-judge, the recently proposed Prometheus 2 \cite{kim2024prometheus2} model is an open-source variant which is specifically trained for qualitative evaluation of model-generated responses and demonstrated higher correlation with humans.

However, research by \cite{wang2023large} and \cite{shen2023large} has highlighted potential limitations in using LLM as evaluators, suggesting that while LLMs can excel in specific areas like translation quality and grammatical error correction \cite{kobayashi2024large,kocmi2023large}, their effectiveness as evaluators may vary significantly across different tasks. Moreover, using closed-source LLMs as evaluators also have associated cost.
This highlights the ongoing debate and research into the capabilities and limitations of LLMs as evaluators in diverse linguistic domains. Therefore, to use LLMs as evaluators, it is important to consider the following:

\begin{itemize}
    \item \textbf{Consistency:} Ensuring consistent combinations of LLMs are used as evaluators when LLMs are used as juries to ensure consistency and reproducibility in assessments.

\item \textbf{Bias and Hallucination Detection:} Developing methods to identify and mitigate bias and hallucinations in the outputs of LLM judges/juries to ensure the reliability and robustness of the evaluation.

\item \textbf{Interpretability:} Enhancing the interpretability of LLM outputs (e.g., asking LLMs to provide reasoning/explanations) to improve understanding and trustworthiness of the evaluation.

\item \textbf{Cost Efficiency:} Advancing the development of efficient LLMs to reduce costs.

\end{itemize}


\if{false}
\begin{figure*}[t!]
    \centering
    \begin{tikzpicture}[
        sibling distance=40mm,
        level distance=50mm,
        edge from parent/.style={draw,-latex},
        every node/.style={font=\small,align=left,text width=30mm}
    ]

    \node [draw, rounded corners, fill=orange!10] (A) {General  Benchmarks} 
        child {node [draw, rounded corners, fill=yellow!20] {
            1. Knowledge Capability and Reasoning: MMLU \cite{hendrycks2020measuring} \\
            2. Common Sense Reasoning: GSM8k \cite{cobbe2021traininggsm8k} , HellaSwag \cite{zellers2019hellaswag}, Winogrande \cite{sakaguchi2021winogrande}, ARC \cite{clark2018thinkarq}, PIQA \cite{bisk2020piqa}, SIQA \cite{sap2019siqa}, OpenBookQA \cite{mihaylov2018openbookqa} \\
            3. Truthfulness: TruthfulQA \cite{lin2021truthfulqa} \\
           4. Coding Capability: HumanEval \cite{chen2021evaluating}, MBPP \cite{austin2021program} \\
        }};
      
    
    \node [draw, rounded corners, fill=orange!10, right=of A] (B) {
    Specific  Benchmarks}
        child {node [draw, rounded corners, fill=yellow!20] {
            1. Format Following Capability: FOFO \cite{xia2024fofo}\\
             2. Conversation Following Capability: MT-Bench \cite{zheng2024judging}\\
            3. Reward Models: RewardBench \cite{lambert2024rewardbench}\\
            4. MultiModal: MMMU \cite{yue2023mmmu}, MathVista \cite{lu2023mathvista}, AI2D \cite{kembhavi2016diagram}, ChartQA \cite{masry2022chartqa}, DOCVQA \cite{mathew2021docvqa} \\ 
        }};

    \node [draw, rounded corners, fill=orange!10, right=of B] (C) { 
   Diverse  Benchmarks}
        child {node [draw, rounded corners, fill=yellow!20] {
            1. LMSys Chatbot Arena \cite{chiang2024chatbot} \\
            2. \href{https://hf.co/open-llm-leaderboard}{OpenLLM} \\
            3. HELM \cite{liang2022holistic}\\
            4. PromptBench \cite{zhu2023promptbench} \\
            5. BigBench \cite{srivastava2022beyond} \\
            6. BigBench-Hard \cite{suzgun2022challengingbigbench} \\
        }};

    \node [draw, rounded corners, fill=orange!10, right=of C] (D) {Independent Benchmarks}
        child {node [draw, rounded corners, fill=yellow!20] {
            1. General Domain: \cite{laskar-etal-2023-systematic}, \cite{bang2023multitaskchatgpt}, \cite{qin2023chatgpt}, \cite{kocon2023chatgptjackofalltrades} \\
            2. Specific Domain: Biomedicine \cite{jahan2024comprehensive}, Finance \cite{guo2023chatgptfinance,li2023largefinance}, Language-specific \cite{lai2023chatgpt,khondaker2023gptaraeval,DBLP:conf/coling/KabirILNBH24}  \\
            3. Ethics, Bias, Toxicity, Trustworthiness: \cite{yang2022gluerobustness,wang2023robustness,zhuo2023redteaming,rawte2023surveyhallucination,liu2023mitigatinghallucination,mcintosh2024inadequacies}.
        }};
    \end{tikzpicture}
    \caption{Taxonomy of Benchmarks.}
    \label{fig:taxonomy}
\end{figure*}
\fi

\if{false}
\subsection{Other things to review}

\begin{itemize}
    \item Model variance leading to 20 point boost on MMLU while not changing performance in other datasets: \url{https://twitter.com/williamwangnlp/status/1773468788958367992}
    \item 
Elo vs MMLU Alignment: \url{https://twitter.com/emollick/status/1787472719065256092}
\item 
Importance of using HELM for LLM Eval: \url{https://twitter.com/percyliang/status/1785878022282965094}
\item Correlations between LLM Benchmarks: \url{https://twitter.com/gblazex/status/1746295870792847562}
\item TIME TRAVEL IN LLMS: TRACING DATA CONTAMINATION IN
LARGE LANGUAGE MODELS: \url{https://arxiv.org/pdf/2308.08493}]
\end{itemize}

\subsection{{Prior Evaluation Papers}}

\begin{itemize}
    \item A Survey on Evaluation of Large Language Models: \url{https://arxiv.org/abs/2307.03109}
    \item Evaluating Large Language Models: A Comprehensive Survey \url{https://arxiv.org/abs/2310.19736}
    \item Lessons from the Trenches on Reproducible Evaluation of Language Models \url{https://arxiv.org/abs/2405.14782}
    \item Leak, Cheat, Repeat: Data Contamination and Evaluation Malpractices in Closed-Source LLMs \url{https://arxiv.org/abs/2402.03927}
\end{itemize}
\fi



\end{document}